\documentclass[runningheads]{llncs}
\usepackage[usenames,dvipsnames]{color}

\usepackage{amsmath}
\usepackage{amssymb}
\usepackage{bm}
\usepackage[linesnumbered,ruled,vlined,resetcount]{algorithm2e}
\usepackage{graphicx}
\usepackage{float}
\usepackage{booktabs}
\usepackage{multirow}
\usepackage[misc,geometry]{ifsym}
\usepackage[hyphens]{url}
\usepackage[hidelinks,breaklinks]{hyperref}
\usepackage{orcidlink}

\newcommand*{\RR}{\mathbb{R}}

\newcommand*{\norm}[1]{\left\lVert#1\right\rVert}

\DeclareMathOperator{\mse}{MSE}

\DeclareMathOperator{\fm}{FM}
\DeclareMathOperator{\im}{IM}
\DeclareMathOperator{\tm}{TM}

\SetAlgoCaptionLayout{footnotesize}
\SetKwComment{Comment}{$\triangleright$\ }{}
\newcommand{\ali}[2]{\makebox[#1][l]{#2}}

\begin{document}

\title{Towards Bio-Inspired Robotic Trajectory Planning via~Self-Supervised RNN}
\titlerunning{Bio-Inspired Robotic Trajectory Planning via~Self-Supervised RNN}
\author{
Miroslav Cibula\inst{1}\textsuperscript{(\Letter)}\orcidlink{0009-0000-8460-6533} \and
Krist\'ina Malinovsk\'a\inst{1}\orcidlink{0000-0001-7638-028X} \and
Matthias Kerzel\inst{2}\orcidlink{0000-0002-1378-0435}
}
\authorrunning{M.~Cibula et al.}

\institute{
Faculty of Mathematics, Physics and Informatics, Comenius University Bratislava, Bratislava, Slovakia \\
\email{cibula25@uniba.sk, kristina.malinovska@fmph.uniba.sk}
\and Department of Informatics, University of Hamburg, Hamburg, Germany \\
\email{matthias.kerzel@uni-hamburg.de}
}

\maketitle
\begin{abstract}

Trajectory planning in robotics is understood as generating a sequence of joint configurations that will lead a robotic agent, or its manipulator, from an initial state to the desired final state, thus completing a manipulation task while considering constraints like robot kinematics and the environment. Typically, this is achieved via sampling-based planners, which are computationally intensive. Recent advances demonstrate that trajectory planning can also be performed by supervised sequence learning of trajectories, often requiring only a single or fixed number of passes through a neural architecture, thus ensuring a bounded computation time. Such fully supervised approaches, however, perform imitation learning; they do not learn based on whether the trajectories can successfully reach a goal, but try to reproduce observed trajectories. In our work, we build on this approach and propose a cognitively inspired self-supervised learning scheme based on a recurrent architecture for building a trajectory model. We evaluate the feasibility of the proposed method on a task of kinematic planning for a robotic arm. The results suggest that the model is able to learn to generate trajectories only using given paired forward and inverse kinematics models, and indicate that this novel method could facilitate planning for more complex manipulation tasks requiring adaptive solutions.

\end{abstract}

\section{Introduction}
\label{sec:intro}
Robotic trajectory planning is vital for robots acting in real-world environments. The robot's planning machinery needs to estimate a viable trajectory for the end effector to reach the goal.
Typically, sampling-based methods are used: random samples are drawn from the robot's configuration space and connected using a local planner. Provided the task goal is specified in Cartesian space, once a trajectory is found, an inverse kinematics (IK) solver computes the corresponding sequence of joint configurations. More advanced approaches refine the trajectory further, aiming for smoothness and compatibility with low-level motor control \cite{coleman2014reducing}.

These methods rely on an accurate model of the robot’s kinematics, which is not always available, especially with custom-built or modified actuators. Moreover, sampling-based methods are generally non-optimal, can have unpredictable planning times, or result in jerky or unnatural movements. This is especially relevant for humanoid robots, where unpredictable or abrupt motions can pose safety risks and hinder human-robot collaboration. Therefore, existing approaches cannot solve all challenges for robots with complex or heavily constrained kinematics, such as humanoids with highly complex hand-like manipulators. 

Developmental robotics \cite{lungarella2003developmental} draws inspiration from biology and cognitive neuroscience: Robots are supposed to learn increasingly complex skills through self-exploration and interaction with the environment in a self-supervised manner, similar to how humans and animals develop motor control \cite{Smith2005}. To follow this paradigm, bio-inspired and neuro-cognitively plausible models for sensory-motor learning are required.

In this proof-of-concept study, we propose a bio-inspired, self-supervised learning scheme for neural trajectory planning in robotics. Our approach proceeds in two stages: first, the robot performs motor babbling to train a forward model (FM) and an inverse model (IM) \cite{Wolpert1998}, both implemented as multilayer perceptrons (MLP) \cite{Cibula2024}. In the second stage, these trained models are embedded within a recurrent neural architecture based on gated recurrent units (GRU) \cite{Cho2014} to reproduce trajectory shapes. By avoiding fully supervised training, which requires large datasets and tends to produce trajectories biased toward the training distribution (even when they fail to reach a goal), we enable the agent to generate novel movements grounded in its interaction with the environment. Although the components are standard neural networks, the overall scheme functionally resembles aspects of human cognition. Thus, our contribution is a bio-inspired architecture that enables a robot to acquire not only kinematics but also movement trajectories from experience, allowing it to generate increasingly complex actions by reusing prior knowledge.
Our results are still preliminary and do not yet provide the full functionality of a trajectory planner; however, they can be a stepping stone towards a biologically plausible and adaptive trajectory planner.

\section{Related Work}

There are several different popular approaches to planning reach and grasp actions. While sampling-based trajectory planners like Probabilistic Roadmap \cite{Kavraki1996} and Rapidly-Exploring Random Tree \cite{LaValle2001} are the de facto standard in applied robotics, 
these planners are not based on biological mechanisms, like learning. Furthermore, if the goal is specified in task space (e.g., Cartesian end-effector space), these approaches require a functioning and fast inverse kinematics solver -- in short, a mathematical model that computes the robot's joint configuration(s) that bring the end-effector into a desired pose (position and rotation). 
The main challenge for inverse modeling remains the cumulative error in the approximated joint position and how to compensate for it \cite{Sinha2019}, as well as the flexibility in trajectory planning to account for obstacles and environmental changes.
One of the recent solutions to this problem in the field of humanoid robotics is the universal neuro-inspired inverse kinematics method, CycleIK, utilizing an MLP with custom $L^1$-based metrics and an additional generative adversarial network for obstacle avoidance \cite{Habekost2024}. 

Inverse models are complemented by forward models \cite{Wolpert1998}, that predict the sensorimotor consequences from a joint configuration sequence, i.e., the next state of the joints and the changes in the observed world. Both of these models 
are usually modeled using various versions of feedforward networks. 
The interconnected forward and inverse models can explain causal learning and mental simulation \cite{Cibula2024}.
Another major approach is the use of reinforcement learning (RL), which learns an action policy without a kinematic model or IK-solver. 
For the review of the RL approaches in the context of this problem, refer to \cite{ElgueaAguinaco2024}. In this work, however, we are focusing on alternative methods that rely on neural network-based sequence modeling task solutions.

Alternatively to the inverse kinematics problem formulation and solutions, planning a trajectory (and related effector action) can be approached with sequence modeling. Sequence modeling in its autoregressive form is a problem considering the inference of a sequence of consecutive states by modeling the state sequence as conditioned on the final desired state.
Typically, this task is solved using recurrent neural networks (RNN) or, more recently, Transformers \cite{Chen2021,Janner2021}. Gated RNNs, such as long short-term memory (LSTM) \cite{Hochreiter1997} or gated recurrent units (GRU) \cite{Cho2014}, have been utilized for robot arm control applications \cite{Chen2022,Otte2016}. Similarly, echo state networks have also been proposed as a solution for movement trajectory prediction \cite{Hellbach2008}, and have recently been enhanced to learn motor primitives for shared control tasks \cite{Amirshirzad2024}. 

\section{Methods}

In our current work, we build on the paired FM and IM as defined in \cite{Cibula2024}. While the FM and IM operate only with time-adjacent states and actions, they cannot be directly used for long-term, multi-step planning. Hence, our proposed approach is to generate trajectories using a sequence model as
\begin{equation}
    \label{eq:tm-def}
    \mathrm{TM} \colon \left[\bm{s}(0), \bm{s}(T)\right] \mapsto \hat{\tau}_s,
\end{equation}
where $\bm{s}(0)$ and $\bm{s}(T)$ denote the initial and goal state in which the task is complete, respectively. The trajectory model (TM) infers an output trajectory of intermediate states $\hat{\tau}_s = \left[\bm{\hat{s}}(1), \bm{\hat{s}}(2), \ldots, \bm{\hat{s}}(T-1)\right]$. Subsequently, it is possible to translate such a trajectory to an action sequence:
\begin{equation}
    \label{eq:act-seq-gen}
    \hat{\tau}_a = \big[ \im\left[\bm{\hat{s}}(t), \bm{\hat{s}}(t+1)\right] \mid 0 \leq t \leq T - 1 \big],
\end{equation}
with $\bm{\hat{s}}(t), \bm{\hat{s}}(t+1) \in \hat{\tau}_s$, and $\bm{\hat{s}}(0) \equiv \bm{s}(0)$, $\bm{\hat{s}}(T) \equiv \bm{s}(T)$.

\begin{figure}[t]
    \centering
    \includegraphics[width=\linewidth]{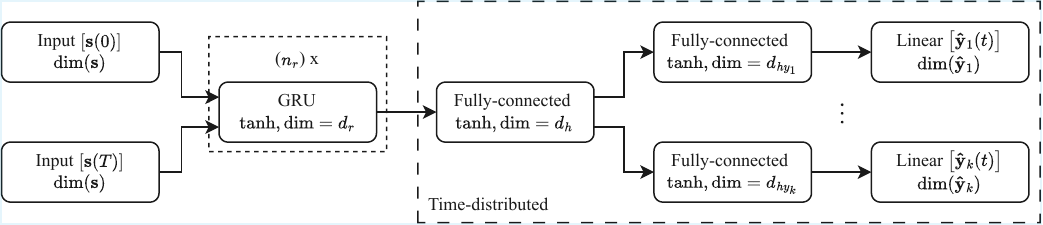}
    \caption{TM architecture implemented as a GRU decoder with $n_r$ $\tanh$-activated GRU layers of $d_r$ units. Input is the initial state $\bm{s}(0)$ and a target state $\bm{s}(T)$. Time-distributed module predicts the sequence of intermediate states' feature components $\bm{\hat{y}}_i (t) \in \bm{\hat{s}}(t)$ for discrete time points $1 < t < T$. Note that the dimensionalities of the hidden layers are task-specific; thus, we do not list them here.}
    \label{fig:tm}
\end{figure}

As a sequence model, the TM was implemented as a recurrent decoder architecture \cite{Cho2014} (Fig.~\ref{fig:tm}) using $n_r$ GRU layers \cite{Cho2014}. Our extensive experimentation \cite{Cibula2024a} has shown that GRU networks have a slight advantage in this context over computationally heavier LSTMs \cite{Hochreiter1997}. In the general architecture, the recurrent layers feed a time-distributed prediction module, which is topologically equivalent to the FM architecture introduced in \cite{Cibula2024}. The module consists of one common fully-connected $\tanh$-activated hidden layer connecting to $k$ output heads, each separately predicting qualitatively different state subvectors $\bm{\hat{y}}_i \subseteq \bm{\hat{s}}(t)$ (e.g., end-effector position, joint configuration, object properties, etc.). Although we use only one output head in our experiment to predict $\bm{{\it ef}}(t)$, the architecture should be able to accommodate more complex state vector prediction using multiple prediction heads.
In our implementation, the TM does not support generating variable-length trajectories. However, it should be possible for the TM to learn to compress the ends of the generated trajectories by producing almost identical states $\bm{\hat{s}}(t)$ if no more steps are needed.

\begin{figure}[t]
    \centering
    \includegraphics[width=0.8\linewidth]{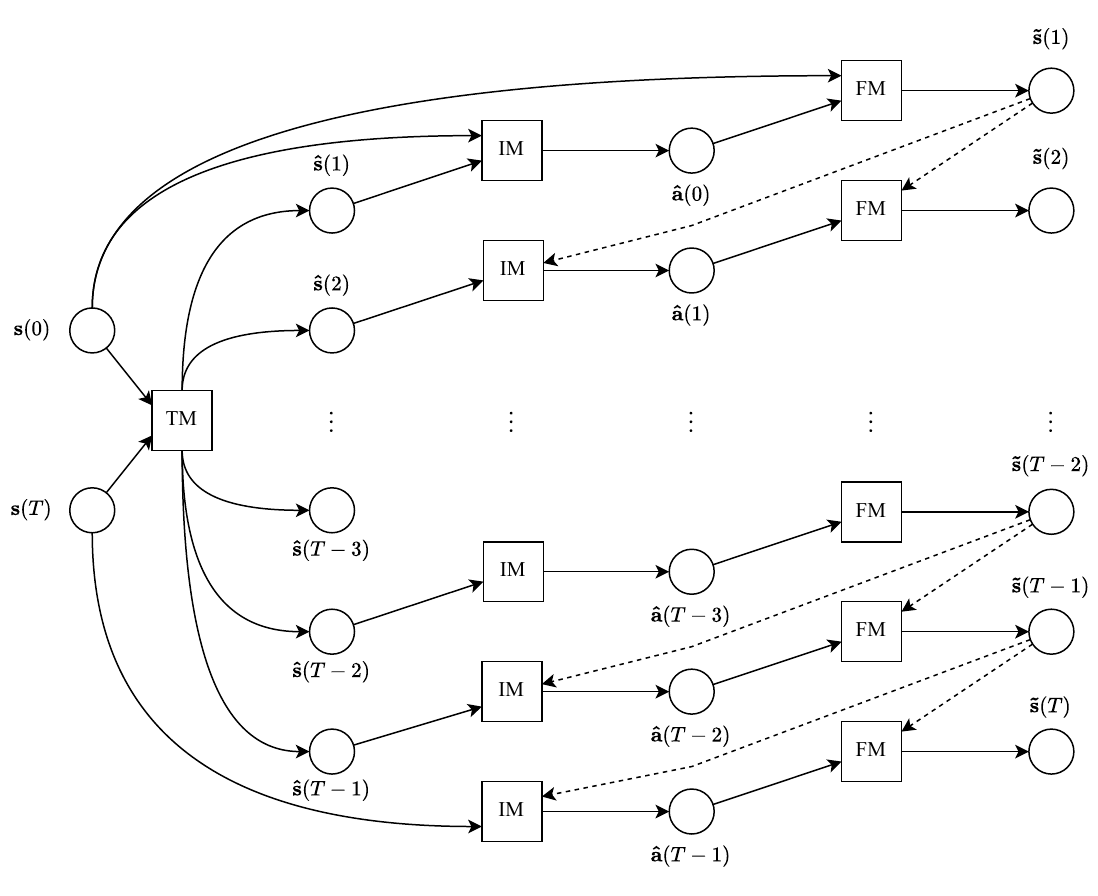}
    \caption{FM/IM-aided generation of the intermediate state trajectory, acting as a partial target for the self-supervised learning of the TM. A \emph{dashed line} designates the flow of the information from the previous timestep contributing to the rectification of the current state.}
    \label{fig:tm-scheme}
\end{figure}

To train the TM, we propose a self-supervised scheme that traverses each generated trajectory and evaluates its states $\bm{\hat{s}}(t) \in \hat{\tau}_s$ in multiple steps (Fig.~\ref{fig:tm-scheme}). The process starts with pairing the first intermediate state $\bm{\hat{s}}(1)$ with the ground-truth initial state $\bm{s}(0)$ and inputting them into the IM inferring action $\bm{\hat{a}}(0)$ responsible for the transition between $\bm{s}(0)$ and $\bm{\hat{s}}(1)$. Using the FM, we then predict the consequence of the execution of $\bm{\hat{a}}(0)$ in $\bm{s}(0)$. This step has the purpose of producing a rectified state $\bm{\tilde{s}}(1)$ with respect to the ground-truth initial state $\bm{s}(0)$. Assuming optimal IM and FM, the FM in this step essentially verifies whether the predicted $\bm{\hat{s}}(1)$ is achievable by the action $\bm{\hat{a}}(0)$ estimated on the basis of $\bm{s}(0)$. If $\bm{\hat{s}}(1)$ is predicted accurately by the TM, discrepancy of $\bm{\tilde{s}}(1)$ against it should be minimal. Thus, we can measure the prediction error of $\bm{\hat{s}}(1)$ as $\mathcal{L} \left[\bm{\hat{s}}(1), \bm{\tilde{s}}(1)\right]$ where
\begin{equation}
    \mathcal{L} \left[\bm{\hat{s}}(t+1), \bm{s}(t+1)\right] \triangleq \frac{1}{k} \sum_{\substack{
        \bm{\hat{y}}_i \subseteq \bm{\hat{s}}(t+1) \\
        \bm{y}_i \subseteq \bm{s}(t+1)
    }} 
        \mse \left(\bm{\hat{y}}_i, \bm{y}_i\right),
\end{equation}
with $\mse$ denoting a mean square error and $k$ being the total number of state feature subvectors.

For the next state $\bm{\hat{s}}(t) \in \hat{\tau}_s$, the process is similar. However, instead of using $\bm{s}(0)$ as the IM and FM input, we utilize the previous rectified state $\bm{\tilde{s}}(t-1)$. This way, we can propagate the correcting signal in the form of more accurate rectified states originating in the ground-truth $\bm{s}(0)$ throughout the whole $\hat{\tau}_s$. This rectified state trajectory generation procedure is formally described in Algorithm~\ref{alg:tm-rect}.

\begin{algorithm}[t]
\DontPrintSemicolon

\BlankLine

\KwIn{Ground-truth initial state $\bm{s}(0)$ and goal state $\bm{s}(T)$}
\KwOut{Trajectory of rectified intermediate states $\tilde{\tau}_s$ and the last rectified state $\bm{\tilde{s}}(T)$}

\BlankLine

\ali{2em}{$\hat{\tau}_s$}      $\gets \tm \left[\bm{s}(0), \bm{s}(T)\right]$ \;
\ali{2em}{$\bm{\tilde{s}}(0)$} $\gets \bm{s}(0)$ \;

\BlankLine

\For{$t = 1, \ldots, T$} {
    \ali{4em}{$\bm{\hat{s}}'(t)$}   $\gets \hat{\tau}_s (t)$ \;

    \BlankLine
    
    \ali{4em}{$\bm{\hat{a}}(t-1)$}  $\gets \im \left[\bm{\tilde{s}}(t-1), \bm{\hat{s}}'(t)\right]$ \;
    \ali{4em}{$\bm{\tilde{s}}(t)$}  $\gets \fm \left[\bm{\tilde{s}}(t-1), \bm{\hat{a}}(t-1)\right]$ \;

    \BlankLine

    \If{$t < T$} {
        \ali{2.34em}{$\tilde{\tau}_s (t)$} $\gets \bm{\tilde{s}}(t)$ \;
    }
}

\caption{Generation of the rectified state trajectory $\tilde{\tau}_s$. $\bm{\hat{s}}'(t)$ denotes $\bm{\hat{s}}(t)$ without its joint configuration subvector $\bm{\hat{\theta}}(t)$ as per the implementation of the IM.}
\label{alg:tm-rect}
\end{algorithm}

With the generated rectified trajectory $\tilde{\tau}_s$, we then compute the prediction error of $\hat{\tau}_s$ as
\begin{equation}
\begin{split}
    \label{eq:tm-loss}
    \mathcal{L}_{\rm TM} \left[\hat{\tau}_s, \tilde{\tau}_s, \bm{s}(0), \bm{s}(T)\right] \triangleq 
    &\frac{1}{\left| \hat{\tau}_s \right|}
        \sum_{i=1}^{\left| \hat{\tau}_s \right|} \mathcal{L} \left[ \bm{\hat{\tau}_s} (i), \bm{\tilde{\tau}_s} (i) \right] \\
    &+ \mathcal{L} \left[\bm{\hat{s}}(1), \bm{s}(0)\right]
    + \mathcal{L} \left[\bm{\hat{s}}(T-1), \bm{s}(T)\right].
\end{split}
\end{equation}
The first term of $\mathcal{L}_{\rm TM}$ measures the average magnitude of discrepancy of predicted trajectory state $\bm{\hat{\tau}_s}(i)$ against the hypothetically more accurate corresponding rectified state $\bm{\tilde{\tau}_s}(i)$, while the subsequent terms compute the distance of the $\hat{\tau}_s$ endpoints to the ground-truth endpoints to position the generated trajectory in space correctly. $\mathcal{L}_{\rm TM}$ is then propagated by standard truncated backpropagation through time \cite{Williams1990}, optimizing the TM.

\section{Experiments}
\label{sec:experiments}

\begin{table}[t]
\centering
\caption{Tested optimizer configurations of the TM training for the kinematics experiment. Each configuration was trained over 50 epochs in 10 trials. $\eta$ denotes an initial learning rate of the Adam \cite{Kingma2014}, RMSprop \cite{Hinton2012}, and stochastic gradient descent (SGD) optimizers. $\gamma$ represents the smoothing constant of RMSprop. Final $\mathcal{L}_{\rm TM}$ is the measurement of $\mathcal{L}_{\rm TM}$ (Eq.~\ref{eq:tm-loss}) from the last epoch of the training.}
\label{tab:traj-opt}
\begin{tabular}{@{}lll@{}}
\toprule
Config. & Optimizer & Final $\mathcal{L}_{\rm TM}$ \\ \midrule
\#1 & Adam($\eta = 10^{-3}$) & $0.084$ \\
\#2 & Adam($\eta = 10^{-4}$) & $0.170$ \\
\#3 & RMSprop($\eta = 10^{-2}, \gamma = 0.99$) & $0.040$ \\
\#4 & RMSprop($\eta = 10^{-3}, \gamma = 0.99$) & $\bf 0.017$ \\
\#5 & SGD($\eta = 0.5$) & $0.034$ \\
\#6 & SGD($\eta = 1.0$) & $0.048$ \\ \bottomrule
\end{tabular}
\end{table}

The feasibility of the proposed method was tested on the task of navigating an end-effector of a 7-DoF robotic arm, KUKA LBR iiwa, from a point in a fixed initial neighborhood in Cartesian space to a random final position, thereby performing trajectory planning. The experiments were performed in a simulated environment using the myGym framework \cite{Vavrecka2021}. A state space consisted of a Cartesian end-effector position $\bm{{\it ef}}(t) \in \RR^3$ and the arm's joint configuration $\bm{\theta}(t) \in \RR^7$, thus $\bm{s}(t) = \left[\bm{\theta}(t), \bm{{\it ef}}(t)\right]$. The action vectors were defined as $\bm{a}(t) = \bm{\theta}(t+1) - \bm{\theta}(t)$, representing change in the joint configuration.
During the simulation, 12,000 trajectories with $T = 11$ (i.e., with 10 intermediate states) were recorded and their endpoints $\left[\bm{s}^{(i)}(0), \bm{s}^{(i)}(T)\right]$ isolated for the training. As is standard in trajectory planning, the goal joint configuration $\bm{\theta}^{(i)}(t)$ was specified at planning time.

Additionally, supplementary FM and IM for this task were separately trained using motor babbling, employing the same architectures and procedures described in Section 4.1 of \cite{Cibula2024}. The FM was trained for 60 epochs using the Adam optimizer \cite{Kingma2014} with an initial learning rate $\eta = 10^{-3}$, resulting in the average mean absolute error (MAE) of the end-effector position and joint configuration prediction of 1.1 mm and $6.3 \times 10^{-4}$ rad. For the IM, the monolithic architecture \cite{Cibula2024} was chosen, and trained for 100 epochs using the AdamW optimizer \cite{Loshchilov2017} with an initial learning rate $\eta = 10^{-3}$ and weight decay $\lambda = 4 \times 10^{-3}$. Its average MAE of action prediction was $1.7 \times 10^{-3}$ rad.

With the trained FM and IM and the collected set of endpoints, we constructed the TM. For this experiment, the architecture of the TM consisted of $n_r = 1$ GRU layer of $d_r = 20$ $\tanh$-activated units feeding the fully-connected common layer of $d_h = d_r$ units. In this instance, there is only one output head predicting $\bm{{\it ef}}$ while $\bm{\theta}$ is computed by the FM. The exclusion of the joint configuration prediction by the TM here is biologically plausible -- when humans perform motor planning, they primarily optimize with respect to task-relevant variables (in this case, the end-effector position) while leaving other variables (such as joint configuration) uncontrolled and conformable to variables of higher priority \cite{Scholz1999}. Topologically, the prediction head in this instance consists of an intermediate hidden layer of $d_{\it hy} = 10$ neurons and outputs to three linearly activated output neurons.

The TM was trained by numerous optimizer configurations with a focus on the Adam \cite{Kingma2014}, RMSprop \cite{Hinton2012}, and stochastic gradient descent (SGD) optimizers (best-performing configurations for each optimizer can be found in Table~\ref{tab:traj-opt}). The training was conducted for 50 epochs on the full dataset of 12,000 endpoints. Afterwards, the TM inferred a trajectory for each pair of endpoints in the dataset. 

\begin{table}[t]
\centering
\caption{Evaluated properties of 12,000 generated trajectories by each TM configuration in Table~\ref{tab:traj-opt}. Init. and final dist. refer to the Euclidean distances between the ground-truth endpoints and the first and last states of generated trajectories, formally $\norm{\bm{s}(0) - \bm{\hat{s}}(1)}_2$ and $\norm{\bm{s}(T) - \bm{\hat{s}}(T-1)}_2$, respectively. Max. sp. dev. represents maximum spacing deviation between the waypoints. All measurements were conducted on whole trajectories, including the initial and final ground-truth states.}
\label{tab:traj-opt-stats}
\begin{tabular}{@{}lllllll@{}}
\toprule
\# & Init. dist. (m)        & Final dist. (m)         & Avg. spacing (m)        & Max. sp. dev. (m)       & Avg. angle ($^\circ$) \\ \midrule
1 & $0.429 \pm 0.064$       & $0.247 \pm 0.066$       & $0.078 \pm 0.014$       & $0.352 \pm 0.054$       & $153.4 \pm 7.4$ \\
2 & $0.632 \pm 0.129$       & $0.296 \pm 0.083$       & $0.110 \pm 0.027$       & $0.522 \pm 0.104$       & $150.3 \pm 4.3$ \\
3 & $0.265 \pm 0.039$       & $0.202 \pm 0.081$       & $0.064 \pm 0.013$       & $0.214 \pm 0.037$       & $142.3 \pm 8.3$ \\
4 & ${\bf 0.134} \pm 0.108$ & ${\bf 0.139} \pm 0.051$ & $0.061 \pm 0.014$       & ${\bf 0.127} \pm 0.083$ & $153.0 \pm 8.8$ \\
5 & $0.183 \pm 0.071$       & $0.239 \pm 0.084$       & $0.063 \pm 0.012$       & $0.194 \pm 0.075$       & $157.5 \pm 9.3$ \\
6 & $0.311 \pm 0.052$       & $0.197 \pm 0.062$       & $0.069 \pm 0.012$       & $0.244 \pm 0.051$       & $156.2 \pm 8.3$ \\ \bottomrule
\end{tabular}
\end{table}

To evaluate the quality of the generated trajectories, we measured several properties. First, we investigated how well trajectories adhere to the ground-truth endpoints. Optimally, for each generated trajectory $\hat{\tau}_s$, its first waypoint $\bm{\hat{s}}(1) \equiv \bm{\hat{\tau}_s}(1)$ must be close enough to the initial point $\bm{s}(0)$ and its last waypoint $\bm{\hat{s}}(10) = \bm{\hat{s}}(T-1)$ to the final point $\bm{s}(11) = \bm{s}(T)$. These criteria are evaluated by computing the Euclidean distance between the given points -- formally, $\norm{\bm{s}(0) - \bm{\hat{s}}(1)}_2$ and $\norm{\bm{s}(T) - \bm{\hat{s}}(T-1)}_2$, respectively, where $\norm{\cdot}_2$ denotes the $L^2$-norm. Furthermore, it is desirable for the TM to produce waypoints as uniformly spaced as possible to achieve movement with a near-constant velocity. For each trajectory, this criterion is evaluated by the average Euclidean distance between every pair of succeeding points and the maximum deviation from this average. Note that the ground-truth endpoints are included in the measurement as well. For optimal waypoint placement, the maximum deviation should be minimal. 

Finally, we measured the angles inside each trajectory to evaluate its smoothness. In the optimal case, we aim to avoid sharp turns within trajectories, as they negatively affect the arm dynamics during execution \cite{Dobis2022}. For each triplet of succeeding waypoints (including the ground-truth endpoints), we measured the angle formed at the middle point. Averaging these measurements across the whole trajectory, the value close to $180^\circ$ indicates an almost linear trajectory, while the lower value suggests that the trajectory is curved. A low minimum angle in the trajectory indicates a sharp turn.

\begin{figure}[t]
    \centering
    \includegraphics[width=0.49\linewidth]{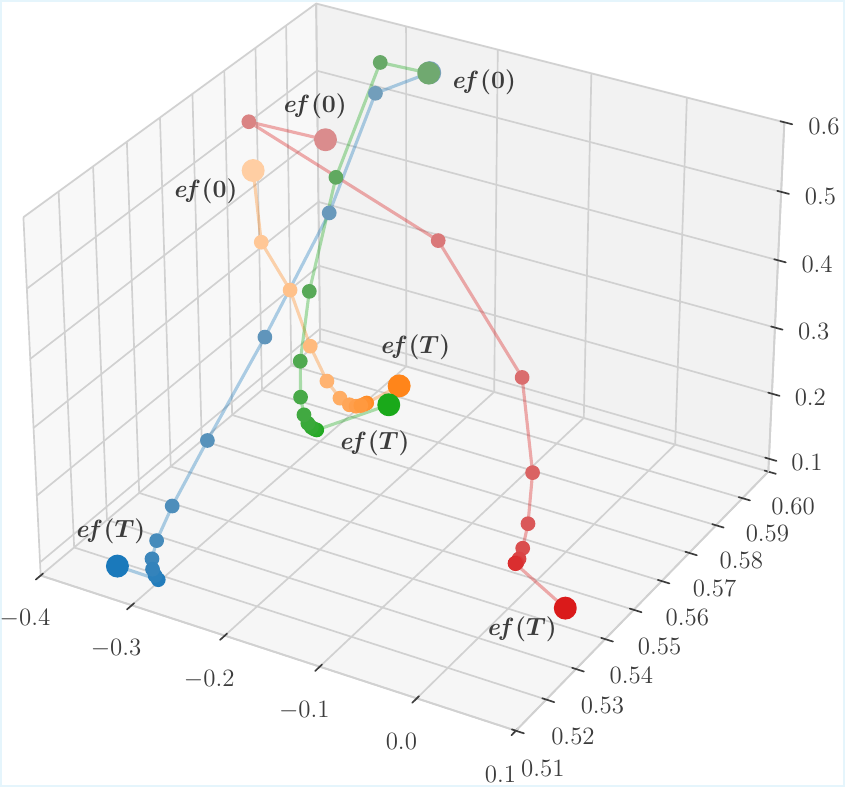}
    \includegraphics[width=0.49\linewidth]{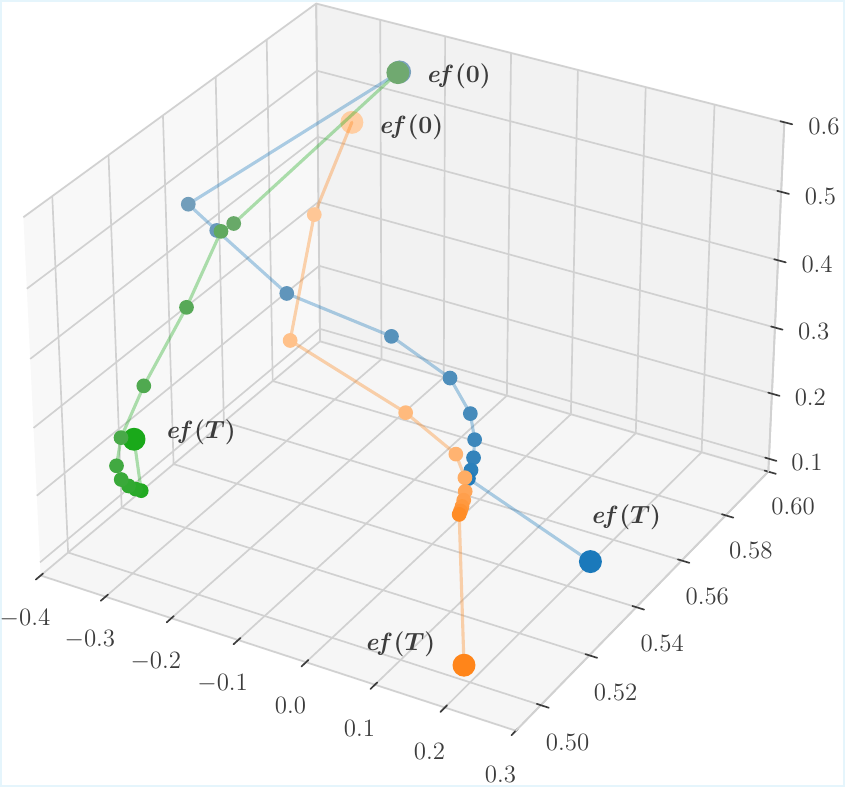}
    \caption{Sample of generated trajectories guiding the arm's end-effector from an initial position ($\bm{{\it ef}}(0)$) to a final position ($\bm{{\it ef}}(T)$). Left: roughly optimal trajectories in reaching target effector positions $\bm{{\it ef}}(T)$. Right: problems with generated trajectories -- uneven spacing between the waypoints; large distances between extreme waypoints.}
    \label{fig:traj-sample}
\end{figure}

The results of the quantitative evaluation of the configurations listed in Table~\ref{tab:traj-opt} can be seen in Table~\ref{tab:traj-opt-stats}. Among the optimizers, any configuration using the Adam optimizer failed to optimize the TM to produce viable trajectories. The most problematic aspect is the distance between the endpoints and the extremes of the trajectories, which reaches, on average, in the best case (config. \#1), $42.9 \pm 6.4$ cm for the initial endpoint and $24.7 \pm 6.6$ cm for the final one. Although such a model is capable of producing more or less smooth trajectories, they are significantly dissociated from the endpoints and thus do not adhere to the task at all. This is in line with the high final $\mathcal{L}_{\rm TM}$ of the Adam models (Table~\ref{tab:traj-opt}). 

\begin{figure}[t]
    \centering
    \includegraphics[width=\linewidth]{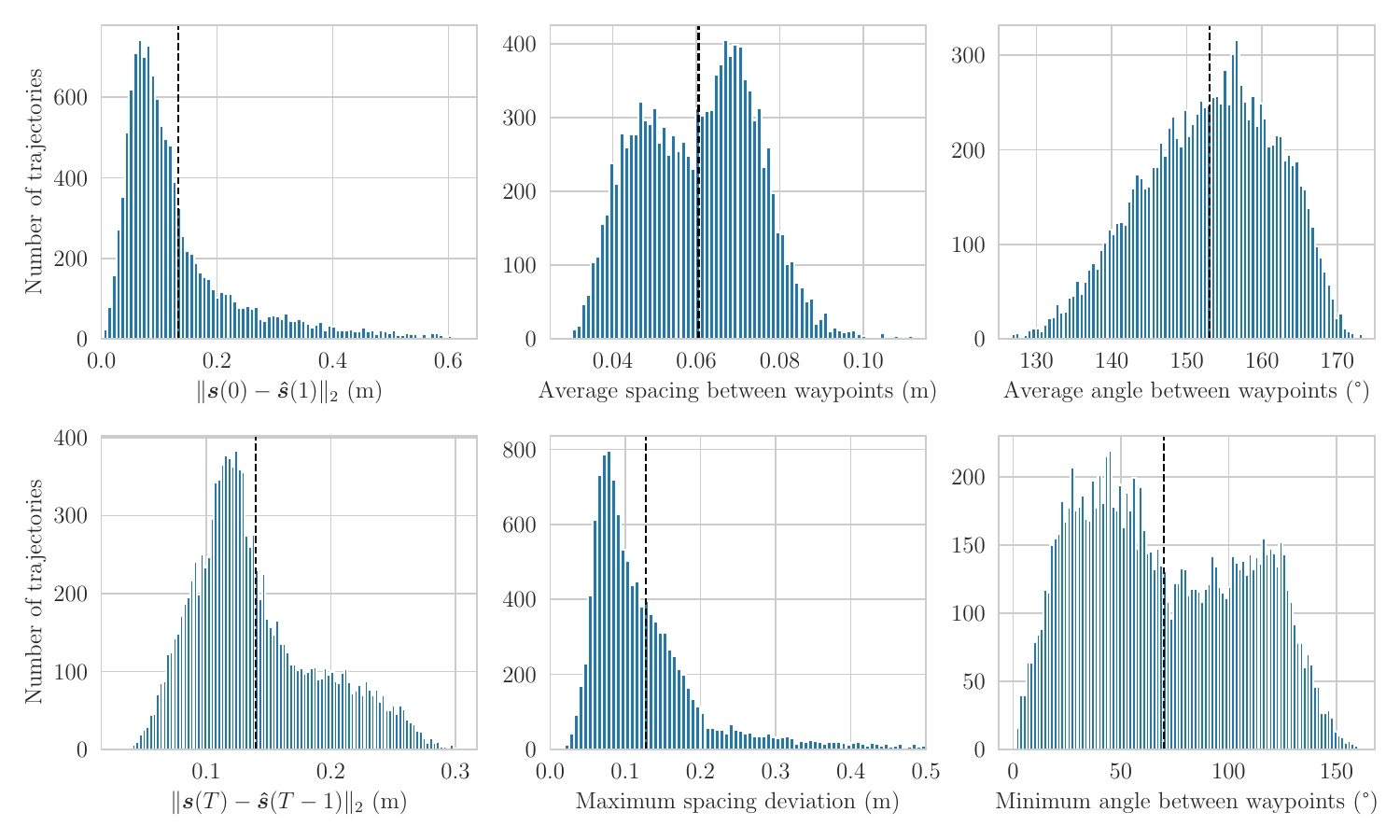}
    \caption{Distributions of measurements over 12,000 trajectories. The first column depicts the distribution of distances between the ground-truth endpoints and the first and last states of generated trajectories. The measurements in the second and third columns were conducted on whole trajectories, including the initial and final ground-truth states. The \emph{dashed lines} represent the expected value of each distribution.}
    \label{fig:traj-hists}
\end{figure}

A similar behavior can be observed in the case of the SGD and RMSprop models with a higher learning rate (configs. \#3 and \#6). Interestingly, still, the TM training seems to require an unusually high learning rate for SGD.
Initially, we hypothesized that the TM would prefer an optimizer with a fixed learning rate (e.g., SGD) over adaptive ones (e.g., RMSprop, Adam). However, results confirmed that the adaptive nature does not pose a problem, as RMSprop generally surpasses SGD in terms of trajectory quality. 

The best configuration, both in terms of trajectory properties and the final $\mathcal{L}_{\rm TM}$, we were able to find was \#4 with the RMSprop optimizer set to a slower learning rate of $\eta = 10^{-3}$. An example of good trajectories produced by this TM can be seen in the left panel of Fig.~\ref{fig:traj-sample}. As shown in Table~\ref{tab:traj-opt-stats} and Fig.~\ref{fig:traj-hists}, on average, the TM was able to position the trajectories with a relatively low distance to the task endpoints; however, these distances still significantly deviate from the average spacing (more than twice on average), which indicates that the trajectory waypoints are usually not well uniformly spread between the endpoints. This property can also be seen in the right panel of Fig.~\ref{fig:traj-sample}.
Although the TM in this configuration generates relatively smo\-oth trajectories with mild curvature (avg. angle $153^\circ$), the left mode of the distribution in the bottom rightmost panel of Fig.~\ref{fig:traj-hists} suggests that 6477 (54\%) generated trajectories contain sharp turns of less than the mean ($69.9^\circ \pm 1.3^\circ$).
Finally, the TM after training in this configuration was able to generate a 10-step trajectory in 993 \textmu s $\pm$ 277 \textmu s (averaged over 12,000 trajectories in 5 trials) running on an Nvidia RTX 4080 SUPER.

\section{Discussion and Future Work}
We present preliminary work towards a biologically inspired neural trajectory planner that is trained in two stages: first, a neural inverse and forward model is trained in a supervised manner via motor babbling. Our evaluation shows that both the FM and the IM are sufficiently accurate (Sect.~\ref{sec:experiments}). In the next step, a recurrent neural architecture utilizing the frozen forward and inverse models is used to generate preliminary trajectories and correct them. We can demonstrate that the approach yields well-positioned trajectories that reach the goal end-effector position on average within a distance of 13.9 cm, while preserving a relatively smooth curve form (Figs.~\ref{fig:traj-sample},~\ref{fig:traj-hists}). In comparison with more trivial methods such as linear interpolation between the endpoints, our approach is able to produce arbitrarily curved trajectories necessary to facilitate more complex planning in the future, for example, for object avoidance.

The proposed model does not require an accurate robotic model (usually provided as a URDF); the required training data can be gained from random movements, both in the form of motor babbling and random trajectories. Moreover, the model incrementally learns more complex tasks that build upon each other, progressing from forward and inverse models to complete trajectories. The model, therefore, fulfills the idea of developmental robotics: learning incrementally more complex abilities from interaction with the environment and the robot's own body.

While our approach could help to gain insight into the development of complex motor-sensory abilities in biological agents, a fully neural trajectory planner is also useful for robotics. In contrast to sampling-based planners, which have unpredictable runtime, our approach produces results within a fixed time. However, we do acknowledge that for application in robotics, some future work must be addressed. First, the trajectory model only learns the robot's kinematics and focuses on the Cartesian end-effector position; it does not account for environmental obstacles and does not directly model joint configurations to verify the kinematic feasibility of the actions produced. In future work, we will evaluate integrating these as an additional input to the neural network. Furthermore, our approach is evaluated on trajectories of comparable length and outputs trajectories with a fixed number of intermediate poses. For more complex and varied trajectories, we will alter the approach to produce trajectories of variable length. We acknowledge that the smoothness and accuracy of the produced trajectories are not competitive with those of optimized trajectory planners. This could be improved by further modifying the loss function of the TM (Eq.~\ref{eq:tm-loss}). For example, by reweighting the last two terms, $ \mathcal{L} \left[\bm{\hat{s}}(1), \bm{s}(0)\right]$ and $\mathcal{L} \left[\bm{\hat{s}}(T-1), \bm{s}(T)\right]$, we suppose the TM could position the extremes of the generated trajectories closer to the ground-truth endpoints, thus increasing accuracy. Finally, in this proof-of-concept study, the proposed approach is tested only on a single task in a simple setting. To further verify the generality of our method, it will be tested on a more diverse and complex set of tasks and environments. An experimental comparison with other related planning methods will also be conducted.

In summary, we contribute a bio-inspired trajectory planner that does not rely on external knowledge about an agent’s kinematics and can thus serve as a model for, e.g., early sensorimotor learning in humans.

\subsubsection{Acknowledgements.} The authors would like to thank Igor Farkaš for his advice and feedback. This research was supported by the Horizon Europe project TERAIS, GA no. 101079338, and in part by the Slovak Grant Agency for Science (VEGA), project 1/0373/23. Research results were partially obtained using the computational resources procured in the national EU-funded project 311070AKF2, National Competence Centre for High Performance Computing.

\bibliographystyle{splncs04}
\bibliography{references}

\begin{thebibliography}{10}
\providecommand{\url}[1]{\texttt{#1}}
\providecommand{\urlprefix}{URL }
\providecommand{\doi}[1]{https://doi.org/#1}

\bibitem{Amirshirzad2024}
Amirshirzad, N., Eren, M.A., Oztop, E.: Context-based echo state networks with
  prediction confidence for human-robot shared control (2024),
  arXiv.2412.00541[cs.RO]

\bibitem{Chen2021}
Chen, L., Lu, K., Rajeswaran, A., Lee, K., Grover, A., Laskin, M., Abbeel, P.,
  Srinivas, A., Mordatch, I.: {Decision Transformer}: {Reinforcement} learning
  via sequence modeling. In: Advances in Neural Information Processing Systems.
  vol.~34, pp. 15084--15097. Curran Associates, Inc. (2021)

\bibitem{Chen2022}
Chen, Z., Walters, J., Xiao, G., Li, S.: An enhanced {GRU} model with
  application to manipulator trajectory tracking. EAI Endorsed Transactions on
  AI and Robotics  \textbf{1},  1--11 (2022). \doi{10.4108/airo.v1i.7}

\bibitem{Cho2014}
Cho, K., van Merri{\"e}nboer, B., Gulcehre, C., Bahdanau, D., Bougares, F.,
  Schwenk, H., Bengio, Y.: Learning phrase representations using {RNN}
  encoder{--}decoder for statistical machine translation. In: Proceedings of
  the 2014 Conference on Empirical Methods in Natural Language Processing
  ({EMNLP}). pp. 1724--1734. Association for Computational Linguistics (2014).
  \doi{10.3115/v1/D14-1179}

\bibitem{Cibula2024a}
Cibula, M.: Cognitively-inspired learning of causal relations in a simulated
  robotic environment. Bachelor's thesis, Comenius University Bratislava (2024)

\bibitem{Cibula2024}
Cibula, M., Kerzel, M., Farkaš, I.: Learning low-level causal relations using
  a simulated robotic arm. In: Artificial Neural Networks and Machine Learning
  -- ICANN 2024. pp. 285--298. Springer Nature Switzerland, Cham (2024).
  \doi{10.1007/978-3-031-72359-9_21}

\bibitem{coleman2014reducing}
Coleman, D., Sucan, I., Chitta, S., Correll, N.: Reducing the barrier to entry
  of complex robotic software: {A} {Moveit!} case study. arXiv preprint
  arXiv:1404.3785  (2014). \doi{10.48550/ARXIV.1404.3785}

\bibitem{Dobis2022}
Dobiš, M., Dekan, M., Beňo, P., Duchoň, F., Babinec, A.: Evaluation criteria
  for trajectories of robotic arms. Robotics  \textbf{11}(1) (2022).
  \doi{10.3390/robotics11010029}

\bibitem{ElgueaAguinaco2024}
Elguea-Aguinaco, {\'{I}}., Inziarte-Hidalgo, I., B{\o}gh, S.,
  Arana-Arexolaleiba, N.: A review on reinforcement learning for motion
  planning of robotic manipulators. International Journal of Intelligent
  Systems  \textbf{2024}(1) (2024). \doi{10.1155/int/1636497}

\bibitem{Habekost2024}
Habekost, J.G., Gäde, C., Allgeuer, P., Wermter, S.: Inverse kinematics for
  neuro-robotic grasping with humanoid embodied agents. In: 2024 IEEE/RSJ
  International Conference on Intelligent Robots and Systems (IROS). pp.
  7315--7322. IEEE (2024). \doi{10.1109/iros58592.2024.10802010}

\bibitem{Hellbach2008}
Hellbach, S., Strauss, S., Eggert, J.P., K{\"o}rner, E., Gross, H.M.: Echo
  state networks for online prediction of movement data -- comparing
  investigations. In: Artificial Neural Networks - ICANN 2008. pp. 710--719.
  Springer Berlin Heidelberg, Berlin, Heidelberg (2008).
  \doi{10.1007/978-3-540-87536-9_73}

\bibitem{Hinton2012}
Hinton, G., Srivastava, N., Swersky, K.: Lecture 6a: {O}verview of mini-batch
  gradient descent (2012),
  \url{https://www.cs.toronto.edu/~tijmen/csc321/slides/lecture_slides_lec6.pdf},
  lecture notes

\bibitem{Hochreiter1997}
Hochreiter, S., Schmidhuber, J.: Long short-term memory. Neural Computation
  \textbf{9}(8),  1735--1780 (1997). \doi{10.1162/neco.1997.9.8.1735}

\bibitem{Janner2021}
Janner, M., Li, Q., Levine, S.: Offline reinforcement learning as one big
  sequence modeling problem. In: Advances in Neural Information Processing
  Systems. vol.~34, pp. 1273--1286. Curran Associates, Inc. (2021)

\bibitem{Kavraki1996}
Kavraki, L.E., Svestka, P., Latombe, J.C., Overmars, M.H.: Probabilistic
  roadmaps for path planning in high-dimensional configuration spaces. IEEE
  Transactions on Robotics and Automation  \textbf{12}(4),  566--580 (1996).
  \doi{10.1109/70.508439}

\bibitem{Kingma2014}
Kingma, D.P., Ba, J.: Adam: {A} method for stochastic optimization  (2014).
  \doi{10.48550/arxiv.1412.6980}

\bibitem{LaValle2001}
LaValle, S.M., Kuffner, J.J.: Randomized kinodynamic planning. The
  International Journal of Robotics Research  \textbf{20}(5),  378--400 (2001).
  \doi{10.1177/02783640122067453}

\bibitem{Loshchilov2017}
Loshchilov, I., Hutter, F.: Decoupled weight decay regularization  (2017).
  \doi{10.48550/arxiv.1711.05101}

\bibitem{lungarella2003developmental}
Lungarella, M., Metta, G., Pfeifer, R., Sandini, G.: Developmental robotics:
  {A} survey. Connection Science  \textbf{15}(4),  151--190 (2003).
  \doi{10.1080/09540090310001655110}

\bibitem{Otte2016}
Otte, S., Zwiener, A., Hanten, R., Zell, A.: Inverse recurrent models -- an
  application scenario for many-joint robot arm control. In: Artificial Neural
  Networks and Machine Learning -- ICANN 2016. pp. 149--157. Springer
  International Publishing. \doi{10.1007/978-3-319-44778-0_18}

\bibitem{Scholz1999}
Scholz, J.P., Schöner, G.: The uncontrolled manifold concept: {Identifying}
  control variables for a functional task. Experimental Brain Research
  \textbf{126}(3),  289--306 (1999). \doi{10.1007/s002210050738}

\bibitem{Sinha2019}
Sinha, A., Chakraborty, N.: Computing robust inverse kinematics under
  uncertainty. In: Volume 5A: 43rd Mechanisms and Robotics Conference.
  IDETC-CIE2019, American Society of Mechanical Engineers (2019).
  \doi{10.1115/detc2019-97945}

\bibitem{Smith2005}
Smith, L., Gasser, M.: The development of embodied cognition: {Six} lessons
  from babies. Artificial Life  \textbf{11}(1–2),  13--29 (2005).
  \doi{10.1162/1064546053278973}

\bibitem{Vavrecka2021}
Vavrečka, M., Sokovnin, N., Mejdrechová, M., Šejnová, G.: {MyGym}:
  {Modular} toolkit for visuomotor robotic tasks. In: 2021 {IEEE} 33rd
  International Conference on Tools with Artificial Intelligence ({ICTAI}). pp.
  279--283. IEEE (2021). \doi{10.1109/ictai52525.2021.00046}

\bibitem{Williams1990}
Williams, R.J., Peng, J.: An efficient gradient-based algorithm for on-line
  training of recurrent network trajectories. Neural Computation
  \textbf{2}(4),  490--501 (1990). \doi{10.1162/neco.1990.2.4.490}

\bibitem{Wolpert1998}
Wolpert, D.M., Kawato, M.: Multiple paired forward and inverse models for motor
  control. Neural Networks  \textbf{11}(7–8),  1317--1329 (1998).
  \doi{10.1016/s0893-6080(98)00066-5}

\end{thebibliography}

\end{document}